\documentclass[10pt,twocolumn,letterpaper]{article}
\usepackage{wacv}
\usepackage{times}
\usepackage{epsfig}
\usepackage{graphicx}
\usepackage{amsmath}
\usepackage{amssymb}
\usepackage{amsmath}
\DeclareMathOperator*{\argmax}{arg\,max}

\wacvfinalcopy % *** Uncomment this line for the final submission

\def\wacvPaperID{495} % *** Enter the wacv Paper ID here

\ifwacvfinal\fi
\begin{document}

%%%%%%%%% TITLE
\title{%Recognizing Object-Related Actions in Real-World Scenarios
 Skeleton-based Action Recognition of People Handling Objects
 % Using  Graph Convolutional Networks}}
 }
%Authors at the same institution
\author{Sunoh Kim$^1$ \hspace{1.5cm} Kimin Yun$^2$ \hspace{1.5cm} Jongyoul Park$^2$ \hspace{1.5cm} Jin Young Choi$^1$ \\
$^1$ASRI, Dept. of Electrical and Computer Eng., Seoul National University, South Korea\\
$^2$Electronics and Telecommunications Research Institute (ETRI), South Korea\\
{\tt\small \{suno8386, jychoi\}@snu.ac.kr, \{kimin.yun, jongyoul\}@etri.re.kr}
}
% Authors at different institutions
% \author{Sunoh Kim \\
% Department of ECE, ASRI
% Seoul National University, Seoul, Korea\\
% {\tt\small suno8386@snu.ac.kr}
% \and
% Kimin Yun \\
% Electronics and Telecommunications Research Institute (ETRI), South Korea\\
% {\tt\small  kimin.yun@etri.re.kr}
% \and
% Jin Young Choi \\
% Department of ECE, ASRI
% Seoul National University, Seoul, Korea\\
% {\tt\small jychoi@snu.ac.kr}
% }

\maketitle
\ifwacvfinal\thispagestyle{empty}\fi

%%%%%%%%% ABSTRACT
\begin{abstract}
In visual surveillance systems, it is necessary to recognize the behavior of people handling objects such as a phone, a cup, or a plastic bag.
In this paper, to address this problem, we propose a new framework for recognizing object-related human actions by graph convolutional networks using human and object poses. 
In this framework, we construct skeletal graphs of reliable human poses by selectively sampling the informative frames in a video, which include human joints with high confidence scores obtained in pose estimation.
The skeletal graphs generated from the sampled frames represent human poses related to the object position in both the spatial and temporal domains, and these graphs are used as inputs to the graph convolutional networks.
Through experiments over an open benchmark and our own data sets, we verify the validity of our framework in that our method outperforms the state-of-the-art method for skeleton-based action recognition. 
\end{abstract}

%%%%%%%%% BODY TEXT
\section{Introduction}
\label{sec:intro}
Human action recognition has attracted extensive attention in recent years, due to its many potential applications in video surveillance, human-robot interaction, and so on. 
Recognizing actions, however, is difficult, not only because of the common challenges of the domain, such as viewpoint changes and occlusions, but also because of the ambiguities of some actions.
% Not only common challenges of the domain such as viewpoint changes and occlusions, but also ambiguities for representing the actions make it harder to recognize actions.
Therefore, human action recognition is still one of the most challenging tasks for computer vision.
% Therefore, human action recognition is still one of the most challenging computer vision tasks.
\begin{figure}[t]
\begin{center}
   \includegraphics[width=1.0\linewidth, bb= 0 0 1500 500]{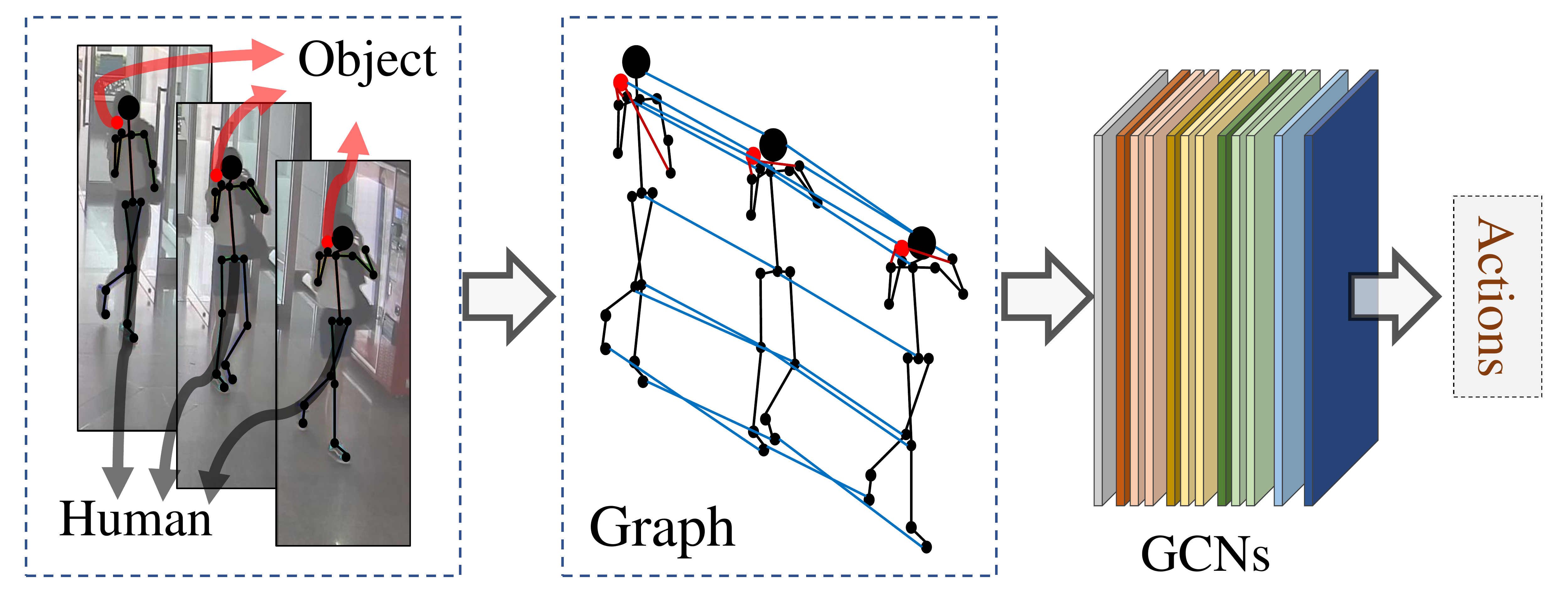}
\end{center}
   \caption{ Action recognition via object-related human poses. 
The pose graph is constructed by using
spatial human poses~(black dots and lines), spatial object poses~(red dots and lines), and temporal connections~(blue lines). 
In spatial and temporal domains, the graph is used as the input to GCNs.
}
\label{fig:gcns}
\end{figure}
% more robust to complex and various backgrounds, as it simplifies a complicated representation of human actions.
% The movements of the body can be interpreted on a physical level through pose
% estimation.
% Among many modalities for human action recognition, such as appearance, optical-flows, depth, and skeletons, pose-based representation using skeletons greatly simplify learning for the recognition, extracting the relevant high-level information.
% \km{위의 어렵다는 말과 연결고리를 만들기. 어렵기 때문에 또한 비디오에서 얻을 수 있는 많은 모달리티를 이용해서 연구들이 진행되고 있다.}
To tackle these challenges in human action recognition, many approaches using various modalities like appearance, optical-flows, depth, and skeleton have been proposed.
Among them, pose-based representation using the skeleton simplifies learning of the actions by extracting the relevant high-level features.
Moreover, the pose-based representation suffers relatively little from the intra-class variances, as the action from actor to actor varies less than in other representations~\cite{yao2012coupled}.
The development of low-cost depth sensors such as Microsoft Kinect and pose estimation algorithms~\cite{cao2017realtime,newell2016stacked,pavlakos2017coarse,shotton2011real} provides an easier way to obtain skeletal information.
It also contributes to the success of data-driven methods of skeleton-based action recognition.

The applications of skeleton-based action recognition in real-world scenarios are difficult due to two major problems.
First, conventional studies~\cite{kim2017interpretable, liu2016spatio,shahroudy2016ntu,stgcn2018aaai} for skeleton-based action recognition usually use constrained datasets, assuming that skeletal data are provided. 
Visual information in real-world scenarios, however, is captured in an unconstrained environment where human skeletons are not given.
Thus, body movement should be interpreted through pose estimation algorithms to obtain skeletal data.
% Thus, pose estimation algorithm is needed to obtain skeletal data for skeleton-based action recognition. 
Pose estimation is a challenging task, however, due to the rarity of appearance or missing or overlapping body parts.
Given that pose estimation results are not always accurate, it becomes crucial to develop approaches dealing with inaccurate poses. % poorly estimated pose
Second, most approaches focusing only on human skeletons ignore contextual information like objects and scenes. 
Most meaningful actions in the real world, though, involve more than one object or person~\cite{moeslund2006survey}.
% \km{which come as the high level representation of human activities~\cite{moeslund2006survey}. }
% \km{윗 문장이 모호함. 다른 말로 쉽게 혹은 삭제. high level representation?}
Therefore, previous methods cannot fully understand the real-world actions.
% because the previous skeleton-based methods remain unable to utilize contextual information.
% \km{문맥이상. 기존방법이 contextual 정보를 사용할 수 없다 -> 사용하지 않기 때문에 이러한 real world 행동의 표현을 잘 할 수 없다라든지.}
% 최근 GCN을 접목한 Skeleton-based action recognition이 성공적인 결과를 내었다. 그 이유는 GCN은 RNN과 CNN과 달리 그래프 자체를 이해하여 정보 손실 없이 high-level feautre를 뽑아내기 때문이다.
% \st{In recent years, graph convolutional networks~(GCNs) have been widely adopted, as they can understand graphs itself by extracting high-level features without loss of graph information. 
% Yan~\textit{et al.}~%\cite{stgcn2018aaai}
% has proposed a spatial-temporal graph convolutional networks~(ST-GCNs) for the skeletion-based action recognition.
% ST-GCN can capture motion information in dynamic skeleton sequences by constructing spatial temporal graphs.
% However, ST-GCN does not consider how to cope with the inaccurate poses and cannot utilize contextual information.} \jy{위 단락의 second, ..의 내용과 중복됨. citation에 포함하고 자세한 내용은 related works에 포함하는 것으로 충분함.}

Motivated by the aforementioned problems, we develop a framework 
for object-related human action recognition based on graph convolutional networks~(OHA-GCN) as depicted in Figure~\ref{fig:gcns}.
To understand the object-related human actions, our method builds the graph structure of both human and object poses.
Human poses are extracted using the pose estimation method~\cite{cao2017realtime}, and object poses are obtained by the pose heatmap and background subtraction.
Using the obtained human and object poses, we model:
% After we obtain human pose and object pose, 
$1)$ the human pose graph, where the movement of human body is considered; 
$2)$ the object-related human pose graph, where the relationship between human and object poses is represented.
Our framework employs the two graphs of human poses and object-related human poses, generated in both the spatial and temporal domains.

To make the graphs more structurally complete, we explore a strategy of selecting informative frames, which discards ambiguous frames.  
After dividing the entire video into segments of equal length, we make a sequence of informative frames by sampling one informative frame from each segment.
Using the confidence scores calculated by the pose estimation algorithm, we can decide which human poses have a more complete structure than the others.
% Then, the two types of the graphs are merged to make pose-based representation for object-related human body.
Then, the two types of the graphs are constructed from the sequence of the informative frames for pose-based representation.
Using the human pose graph and object-related human pose graph, OHA-GCN runs in the architecture of the two stream to boost action recognition.
The GCNs in each stream of our framework apply convolution filters to each graph.
The nodes and their neighbors in each graph are applied to the convolutional filters. 
In experiments, our framework is validated using an open benchmark and our own data sets that includes an illegal rubbish dumping~(IRD) dataset, which compiles realistic data in unconstrained environments.

\section{Related works}
%-------------------------------------------------------------------------
\subsection{Graph convolutional neural networks}
% \km{위 section 뒷부분에 나오는 GCN과 같은 내용이면 위의 section에서 GCN 내용을 아래로 가져오기}
% \kso{skeleton-based 에서 GCN이 갑자기 튀어나오는거처럼 보일 수 있으니 GCN을 related work 앞 섹션으로 옮기는게 나아보여서 옮겼습니다. 그리고 형 말씀대로 skeleton-based섹션에서 GCN이 겹치긴 하는데,  skeleton-based 섹션에서 쓴 GCN과 CNN, RNN은 skeleton-based methods 중 하나로서 human pose를 예시로 들며 썼는데 여기로 옮기면 약간 안맞을 수도 있다고 생각했는데, 아니면 어떻게 옮기면 좋다고 생각하시나요?!}
% \km{괜찮은 것 같기도. 한번 읽어보고 알려줄께. 일반적인 GCN 설명 부분이니 여기있어도 될 것 같고}

Recent advancements in deep neural networks have led to the development of graph convolutional networks~(GCNs) to understand the form of graph structures~\cite{bruna2013spectral,defferrard2016convolutional,duvenaud2015convolutional, kipf2016semi, li2015gated,niepert2016learning}. 
GCNs generalize convolutional neural networks~(CNNs) from low-dimensional grids of images to high-dimensional domains represented by arbitrarily structured graphs.
These tasks are categorized into two main categories: spectral perspective and spatial perspective methods. 
Spectral perspective methods convert graph data into a spectrum and apply CNNs to the spectral domain~\cite{defferrard2016convolutional,duvenaud2015convolutional, li2015gated}.
Different from the spectral perspective methods, spatial perspective methods directly use graph convolutions to define parameterized filters~\cite{bruna2013spectral,niepert2016learning}.
The convolution operation in the spatial perspective resembles the convolution operation on images.
We propose a model using GCNs for action recognition following the concepts of the spatial perspective.

%-------------------------------------------------------------------------
\subsection{Action recognition}
Video-based action recognition is a challenging task because the concepts of the actions are highly abstract and consider both spatial and temporal dimensions.
Conventional deep-learning-based approaches for action recognition rely mainly on 3D CNNs and two-stream CNNs.
The methods based on the 3D CNNs propose a model that directly applies 3D convolution filters to RGB video sequences~\cite{carreira2017quo,ji20133d, tran2015learning}.
3D CNNs, however, have problems in training due to the explosion of parameters, and their performance is only marginally improved compared to the traditional method~\cite{wang2013action}.
Methods based on the two-stream architecture of CNNs are proposed to solve these problems~\cite{feichtenhofer2016convolutional,simonyan2014two, wang2016temporal}.
In these methods, two CNNs are applied to process appearance and motion~(optical flows) independently.
These methods outperform traditional methods using hand-crafted features~\cite{dalal2005histograms, laptev2008learning, wang2013action}.
But, the heavy computation requirements for optical flows become a main limitation of the methods.
Our approach is based only on pose information that can be extracted from RGB images. 
The pose-based representation can reduce computation time and simplify training for the recognition.

%%%%%%%%%%%%%%%%%%%%%%%%%%
\begin{figure*}
\begin{center}
% \fbox{\rule{0pt}{2in} \rule{.9\linewidth}{0pt}}
\includegraphics[width = 1.0 \linewidth, bb= 0 0 950 250]{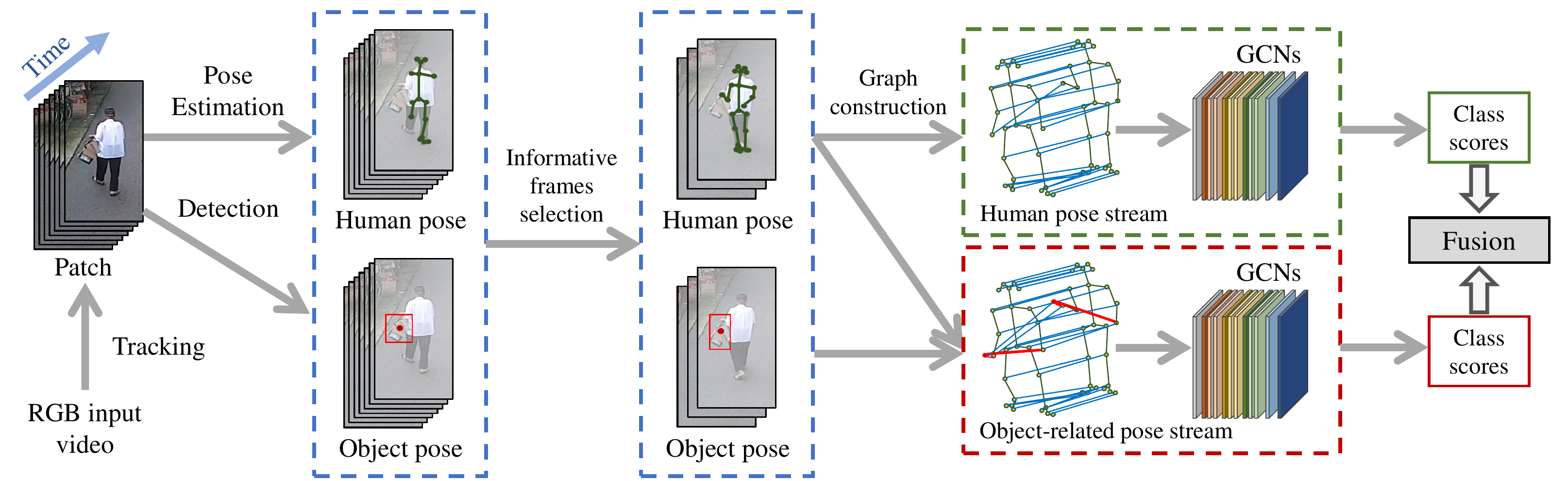}
\end{center}
   \caption{The overall scheme of the proposed framework that includes pipeline modules of tracking and detection for extracting informative human and object poses, pose estimation, graph construction of human and object-related human poses, and GCNs for recognition of object-related human actions.} 
\label{fig:pipeline}
\end{figure*}
%%%%%%%%%%%%%%%%%%%%%%%%%%
%-------------------------------------------------------------------------
\subsection{Skeleton-based action recognition}
Some methods for action recognition are based on skeletal data because human poses are highly relevant to human action.
The methods can be categorized mainly into those based on hand-crafted features and those based on deep-learning features.
The former methods design several features for understanding human joint motions~\cite{fernando2015modeling, vemulapalli2014human, wang2016graph}. 
For example, the human skeleton is represented as a point in the Lie group~\cite{vemulapalli2014human}, encompassing rotations and translations between body parts.
With the recent success of deep learning, many methods for skeleton-based action recognition have been proposed.
Deep-learning-based methods usually use well-established neural models like recurrent neural networks~(RNNs)~\cite{liu2016spatio,shahroudy2016ntu,zhang2017geometric} or CNNs~\cite{ke2017new,kim2017interpretable}.
RNNs use their internal memory to process sequences of input data, making them applicable to skeleton-based action recognition.
Zhang~\textit{et al.}~\cite{zhang2017geometric} select geometric features based on distances between human joints and use an RNN-based model.
CNN-based methods propose applying the CNN directly by considering the skeleton sequence as an image array.
% CNN-based methods are also proposed because CNNs are able to learn not only 2D arrays, but also a spatial temporal representation like skeleton sequences.
Qiuhong~\textit{et al.}~\cite{ke2017new} transform the skeleton sequence into three clips and learn the spatio-temporal feature based on deep CNNs.
Still, generalizing the CNNs and RNNs to work on graph structures is challenging because they cannot fully understand the graphs. 
Since human poses are represented by a graph structure physically, the methods based on the CNNs and RNNs are limited in their ability to use the poses.
Recently, the GCN-based methods have been suggested to understand the human poses without loss of graph information~\cite{tang2018deep,stgcn2018aaai}.
Yan~\textit{et al.}~\cite{stgcn2018aaai} propose a ST-GCN that applies the GCNs for skeleton-based action recognition.
This method can capture the motion information in dynamic skeleton sequences by constructing spatio-temporal graphs.
However, ST-GCN does not consider how to cope with inaccurate poses and cannot understand contextual information.
% However, ST-GCN is unable to understand the contextual information, and can not construct structurally complete graphs of human pose.
By contrast, our method can build the pose graphs with the contextual information using the reliable poses through the informative joint selection.

%-------------------------------------------------------------------------

\section{Object-related  human  action recognition}

\subsection{Overall scheme}
\label{sec:overall}
The overall scheme of the proposed framework is depicted in Figure~\ref{fig:pipeline}. Pose estimation and detection algorithm are applied to patches extracted from RGB video through a tracking algorithm. 
Then, we make a sequence of informative frames by selecting frames containing relatively accurate poses in the patches. 
Using a spatial-temporal graph of human poses and object-related human poses from the informative frames, the GCNs perform action recognition
in the architecture of the two stream. 
We revisit the ST-GCN~\cite{stgcn2018aaai} in Section~\ref{sec:gcn-review}.
We then present how our method detects the objects handled by humans in Section~\ref{sec:obj-detect} and give detailed descriptions of the object-related human action recognition~(OHA-GCN) framework.
Section~\ref{sec:graph-construct} and Section~\ref{sec:GCNs} explain how to construct graphs of object-related human poses and apply the graphs on the GCNs, respectively.
% How to construct graphs of object-related human poses and apply the graphs on the GCNs are explained in Section~\ref{sec:graph-construct} and Section~\ref{sec:GCNs}, respectively.
Finally, we present the strategy of selecting informative frames in Section~\ref{sec:inf-select}.
%-------------------------------------------------------------------------
\subsection{ST-GCN \cite{stgcn2018aaai}}
\label{sec:gcn-review}
ST-GCN uses a spatial temporal graph to form hierarchical representation of the human skeleton. 
The spatial temporal graph consists of nodes and edges where each node corresponds to a joint of the human body, and each edge corresponds to the spatio-temporal connection of the node.
In one single frame, they formulate a spatial graph which is represented by human body joints, and the spatial edge of the graph becomes natural connection between human body joints.
Then, temporal edges are made by connecting the corresponding joints in consecutive frames.

The skeletal graph with $N$ joints in one single frame and $T$ frames of a video is represented by $G=(V,E)$, where $V=\{v_{ti}|t=1,...,T,\; i=1,...,N\}$ is the node set.
In terms of spatial domain, the graph convolution is written as 
\begin{equation}
\begin{aligned}
\label{eq:graph_conv}
  f_{out}(v_{ti})=\underset{v_{tj} \in B(v_{ti})}{\sum}\frac{1}{Z_{ti}(v_{tj})}f_{in}(v_{tj})\cdot \mathbf{w}(l_{ti}(v_{tj})),
\end{aligned}
\end{equation}
where $f_{in}$ and $f_{out}$ are the input and output feature map, respectively. 
$B(v_{ti})$ is the neighbor set of a node $v_{ti}$, where the 1-distance neighbors of the target node $v_{ti}$ are considered.
$\mathbf{w}$ is the weight function which provides a weight vector to compute inner product with the input feature map.
Because the number of the weight vectors is fixed, strategies to partition $B(v_{ti})$ into a fixed number of subsets have to be designed. 
$l_{ti}$ is a function mapping each node in neighborhood of $v_{ti}$ to its subset label. 
To prevent the different subsets from unbalancing the output, the cardinality of the subset $Z_{ti}(v_{tj})$ is used as the normalizing term.
In ST-GCN, many partitioning strategies are introduced, such as uni-labeling, distance partitioning, and spatial configuration.
Among them, spatial configuration has achieved best performance, as it consider the gravity center of the human body.
This strategy divide the neighboring nodes of target node into three subsets: 
$1)$ the target node itself; 
$2)$ centripetal subset: the nodes that are closer to the gravity center; 
$3)$ otherwise centrifugal subset.

The implementation of the graph convolution~\cite{kipf2016semi} is adopted for ST-GCN. 
$\mathbf{A}$ is the $N\times N$ adjacency matrix where a graph structure is represented as matrix form by indicating connections between nodes.
% $\mathbf{I}$ is the $N\times N$ identity matrix which represents self-connection of nodes.
To implement ST-GCN with the spatial configuration partitioning, the Eq.(~\ref{eq:graph_conv}) is transformed into
\begin{equation}
\begin{aligned}
\label{eq:graph_for_spatial}
  \mathbf{f}_{out}=\overset{}{\underset{j}{\sum}}\mathbf{\Lambda}_j^{-\frac{1}{2}}\mathbf{A}_j\mathbf{\Lambda}_j^{-\frac{1}{2}}\mathbf{f}_{in} \mathbf{W}_j,
\end{aligned}
\end{equation}
where the input feature map $\mathbf{f}_{in}$ is represented as a tensor $(N, T, C)$ dimensions, $C$ being the number of input channels.
% With spatial configuration partitioning, the kernel size $K$ is $3$.
Then, the adjacency matrix is divided into three matrices: 
$1)$ $\mathbf{A}_0$, the self-connection of each node; 
% $1)$ $\mathbf{A}_0$, the self-connection of each node, also written as $\mathbf{A}_0=\mathbf{I}$; 
$2)$ $\mathbf{A}_1$, the connections of the centripetal subset; 
$3)$ $\mathbf{A}_2$, the connections of the centrifugal subset.
$\mathbf{W}_j$ denotes the weight matrix, where weight vectors of multiple output channels are stacked.
$\mathbf{\Lambda}_j$ whose element can be formulated as $\Lambda^{ii}_j=\sum_k(\mathbf{A}_j^{ik})+\alpha$ is the diagonal node degree matrix of $\mathbf{A}_j$. 
$\alpha$ is set to 0.001 to avoid empty rows in $\mathbf{A}_j$.
% Also, ST-GCN implements the learnable edge importance weighting, adding a $N\times N$ attention map to Eq.~\ref{eq:graph_for_spatial}.
% But we skip it here for better understanding.

% ST-GCN is not suitable for real-world applications due to the graph structure of inaccurate pose and inability to utilize contextual information.
% Furthermore, ST-GCN only captures human pose information, ignoring object information which is an important clue for recognizing the human-object interactions.
%-------------------------------------------------------------------------
\subsection{Detection of objects handled by humans}
\label{sec:obj-detect}

ST-GCN can capture dynamics of human poses by learning both the spatial and temporal patterns of skeletal data. 
ST-GCN has limitations, however, in understanding object-related human actions because it focuses on only the human poses.
For example, the object-related actions, such as dumping, phoning, texting, and smoking require a specific object that acts as a main cue for recognition.
Inspired by the issue, we create a new graph structure to form a hierarchical representation of the human and object poses.
Before constructing the graphs, it is essential to detect the objects a human is handling.

% In order to apply the IPN to applications such as CCTV camera, it is essential to detect objects that human is carrying.
An easy approach is to find objects that overlap with humans through the object detector, but this approach has several problems.
First, objects in the background that are not related to humans, such as parked cars, are regarded as related objects.
To solve this problem, we define a class of objects handled by humans and develop a new detector, but the shapes of the objects handled by humans vary widely.
The objects handled by humans have a variety of shapes and are small compared to humans. 
Hence, conventional detectors cannot easily detect the objects handled by humans.
Several methods have been presented to use detection networks to detect humans, objects, and their interaction~\cite{Hu:2018vj,Gkioxari:2018uv}. 
Still, a pair consisting of a human and an object is required to train the model on each action, and the training operates independently for each frame, not for the whole video.

In this paper, we use a simple and efficient method to find an object handled by a human through the temporal property of a video.
Since CCTV cameras are generally fixed and limited to the motions of Pan-Tilt-Zoom, it is possible to find a moving area effectively through background modeling and camera motion compensation~\cite{Kim:2012bo,Yi:2013gp,Yun:2017jd}.
This moving area includes the areas of both the human and the object carried by the humans. 
Then, object area can then be detected by subtracting the human area from the moving area regardless of the shape of the object. Fortunately, a human area can be obtained by using Openpose, a 2D pose estimation algorithm providing the location of each joint. The algorithm provides not only the location but also the confidence scores of the joints and affinities between joints as a heatmap, which can then be used as a human area.

\begin{figure}[t]
\begin{center}
   \includegraphics[width=1.0\linewidth, bb= 0 0 700 500]{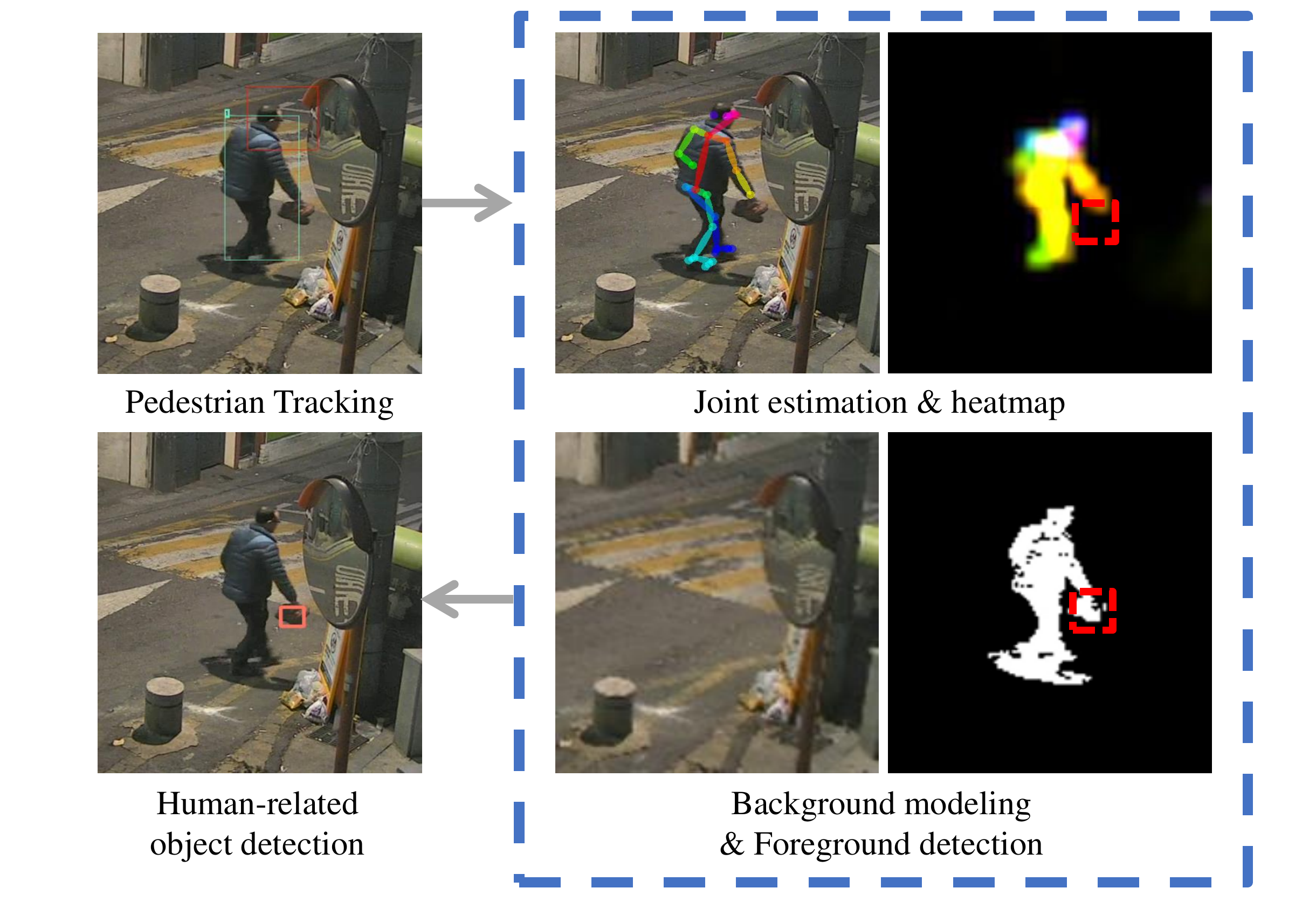}
\end{center}
   \caption{The procedure for detecting the object handled by a human. A moving area is obtained by background subtraction with camera motion compensation and the human area is found by estimated joint and heatmap. Then, the object area is detected by subtracting the human area from the moving area.
   }
\label{fig:object-det}
\end{figure}

Figure~\ref{fig:object-det} shows the procedure for detecting an object handled by a human.
In order to detect the moving region, we use the ViBE algorithm~\cite{Barnich:2011ks} which efficiently builds the background model by storing a set of pixel values from the past.
Then, through the $Openpose$ algorithm, we get each coordinate of the joint and the joint heatmaps.
We aggregate the heatmap for all joints and all joint association into a human region map, which is subtracted from the moving area.
The regions around each human hand are selected.
Finally, we can construct a graph structure by connecting the center position of this region with the joints of the closest person.

%-------------------------------------------------------------------------
\subsection{Graph for object-related human poses}
\label{sec:graph-construct}
We use our detection algorithm and $Openpose$ to obtain the coordinates of the human joints and objects for creating a graph structure of the human and object poses.
% the $Openpose$ algorithm is used.
For human skeletal data, the pose estimation algorithm produces 25 joints in a single frame.
We then formulate a human pose graph using all the human joints in a video sequence.
In addition, the object position obtained in Section~\ref{sec:obj-detect} is added to the human pose graph for modeling our object-related human pose graph.
As a result, the object-related human pose graph consists of 26 nodes that represent human and object poses.
We can model a spatial graph of the physical connection between joints in the human body and position of the object in a frame.
The corresponding nodes in consecutive frames are then connected to formulate a temporal graph.
% Before constructing object-related human pose graph, we use the pose estimation and detection. 

\begin{figure}[t]
\begin{center}
   \includegraphics[width=1.0\linewidth, bb= 0 0 800 500]{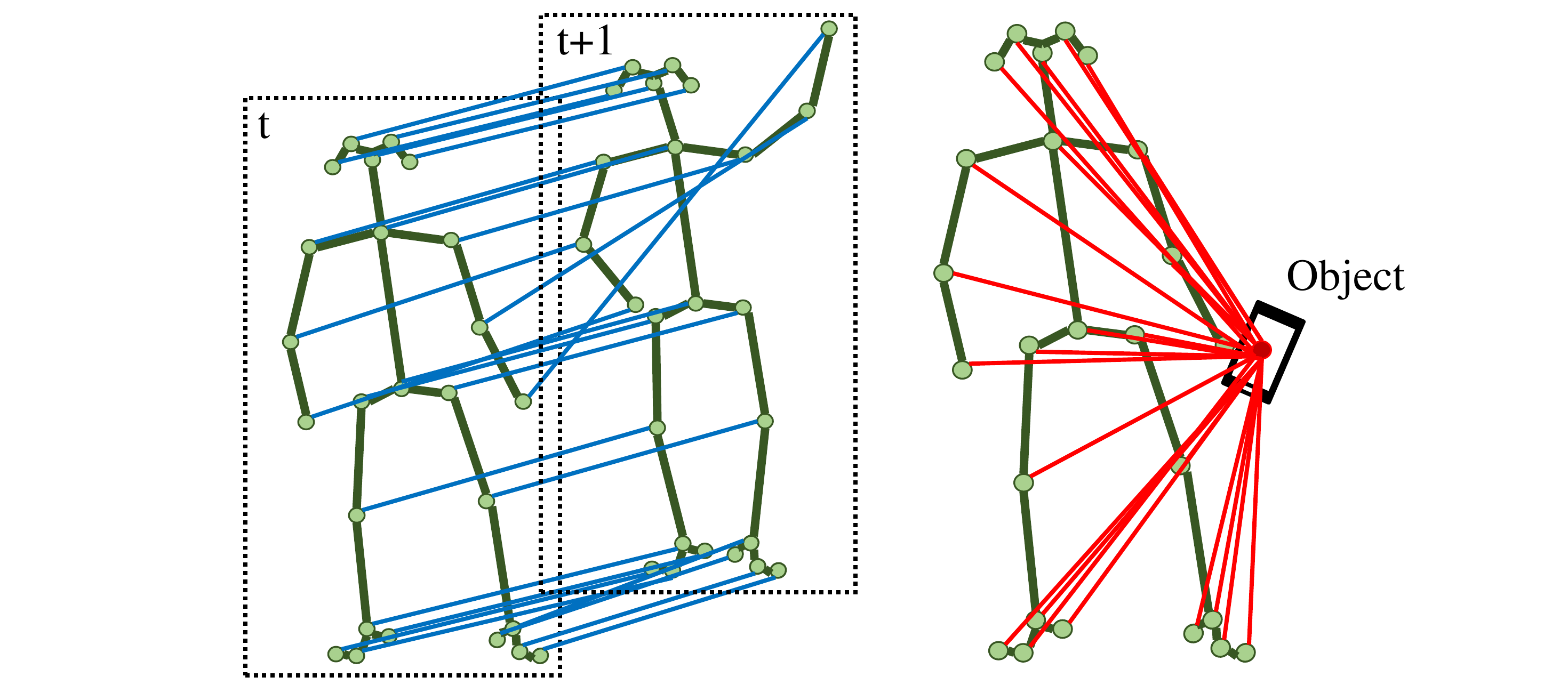}
\end{center}
   \caption{ For the construction of our skeletal graph, all of the dots are used as nodes and all of the lines as edges. In each frame, nodes and edges based on the natural human body are denoted as green.   \textbf{Left:} The graph structure of human poses in terms of both the spatial and temporal domains. Blue lines represent the temporal connections of corresponding nodes in consecutive frames.
   \textbf{Right:} The example of the graph structure for object-related human poses in each frame. 
   A red dot denotes the position of a human-related object. 
   Red lines are the connections between all human body joints and the object.
   }
\label{fig:skeleton_graph}
\end{figure}

To avoid confusion, we use notation similar to that used in the ST-GCN.
In our skeletal graph $G=(V,E)$ for object-related human poses, the node set is denoted as $V=\{v_{ti}|t=1,...,T,\; i=1,...,N\}$ with $N$ joints in one single frame. 
$T$ is the number of informative frames sampled from the video.
The details of the informative frames are explained in Section~\ref{sec:inf-select}.
The edge set is $E=E_S\cup E_T$, which depicts two subsets for spatial and temporal connection.
The subset for spatial connection is 
\begin{equation}
\begin{aligned}
\label{eq:graph_const}
  E_S=\{e_{tij}|v_{ti},v_{tj}~\textit{are connected}, (i,j)\in O\},
\end{aligned}
\end{equation}
where $O$ is the set of the connections for natural human body and the connections between an object and the human body joints.
The subset for temporal connection
\begin{equation}
\begin{aligned}
\label{eq:graph_const2}
    E_T=\{e_{ti}|v_{ti},v_{(t+1)i}~\textit{are connected}\}
\end{aligned}
\end{equation}
% $E_T=\{e_{ti}|v_{ti},v_{(t+1)i}$ \textit{are connected}$\}$
connects the same nodes which is temporally consecutive.
An example of our spatial-temporal skeleton graph is shown in Figure~\ref{fig:skeleton_graph}.

We consider how to connect the object with the human body.
Many strategies of the connection are explored empirically, but we discuss three strategies which are representative of many cases in our view.
First, we use the relationship between the whole joint and the object.
This method, however, makes the graph-based network inefficient because the graph includes redundant information.
Thus, we build a graph structure that is denoted as limbs, focusing on the specific body parts used for interactions. 
We also introduce a strategy using only human hands, where most object-related human actions occur.
The three strategies are denoted as follows: 
$1)$ Body: connections between all the human joints and the object; 
$2)$ Limbs: connections between the human joints belonging to arms or legs and the object; 
$3)$ Hands: connections between the human joints of both hands and the object.
Detailed experiments with each strategy are described in Section~\ref{sec:ablation}.

% \begin{figure*}
% \begin{center}
% % \fbox{\rule{0pt}{2in} \rule{.9\linewidth}{0pt}}
% \includegraphics[width = 1.0 \linewidth]{figure/pipeline.pdf}
% \end{center}
%   \caption{The overall pipeline of our proposed method.
%   Our method applies the pose estimation and detection algorithm to patches extracted from RGB video through tracking algorithm. 
%   Then, we make a sequence of informative frames, selecting frames of relatively accurate poses in the patches. 
%   Using a spatial temporal graph of human poses and object-related human poses from the informative frames, the GCNs perform action recognition.}
% \label{fig:pipeline.}
% \end{figure*}

%-------------------------------------------------------------------------
\subsection{Two stream GCNs with human and object poses}
\label{sec:GCNs}
% \subsection{Graph convolutional networks using human and object pose}
% \km{section 이름 간결하게 해서 한줄안에?}
% \kso{제목이 한 줄이 되도록 수정.}
Using only the human pose graph is limited to express only human body motion, ignoring human-object interactions.
Based on the graph structure of both human and object-related human poses, we propose a framework of two-stream GCNs called OHA-GCN for object-related human action recognition.
The OHA-GCN can boost the recognition for object-related actions, as the human poses and object-related human poses are jointly used.
Human-related objects do not, however, appear in every scene. 
In this regard, the two-stream GCN is designed to use the object-related human pose stream only if human-related objects are found.
In the absence of the objects, OHA-GCN uses only the human pose stream. % for the recognition. 
If the object is properly detected, OHA-GCN recognizes actions through both the human pose stream and the object-related human pose stream.
% OHA-GCN runs in the architecture of the two-stream.

The GCN in each stream takes the graph constructed in Section~\ref{sec:graph-construct} as input and produces a video-level prediction for action recognition.
The model of GCN is composed of 9 GCN layers. Each layer has 64, 64, 64, 128, 128, 128, 256, 256, and 256 channels, respectively.
The global average pooling layer then follows and produces the feature vector. The final outputs are produced by Softmax for action prediction.
Categorical cross-entropy loss is adopted for training of the GCN.
We adopt evenly averaging Softmax scores for final predictions through empirical evaluation of different aggregation methods such as averaging and maximum.

%-------------------------------------------------------------------------
\subsection{Informative frame selection strategy}
\label{sec:inf-select}

The conventional skeleton-based action recognition methods assume that accurate human skeletons are provided by datasets. 
Real-world scenarios, however, do not provide skeletal data, and the estimated skeletal data might be inaccurate due to an unusual appearance or missing or overlapping body parts.
Likewise, since our methods are based only on RGB sequences, the pose estimation results may be inaccurate.
This issue motivates us to introduce a new sampling strategy that selects informative frames. 
Our informative frame selection strategy extracts a short sequence of sampled frames from a video sequence. 
The sampled frames preserve the joints enough to understand human poses with high confidence score from the $Openpose$ algorithm.
Based on confidence scores, we selectively discard frames containing ambiguous human poses.
Then our method can select only those sequences with reliable human pose information. 
%to perform video-level prediction.

Initially, the entire video is divided into $T$ segments $\{S_1,S_2,...,S_T\}$ at equal interval.
We construct sequences of $T$ frames by sampling one frame from each segment.
Since consecutive frames have little variation and provide redundant pose information, the sampled sequences can reduce the redundancy.
Each segment has $M$ frames, so $t$-th segment $S_t$ has frames $\{F^1_{t},F^2_{t},...,F^M_{t}\}$, where $F^m_{t}$ is $m$-th frame in the $t$-th segment.After applying the pose estimation algorithm to each frame $F^m_{t}$, we can obtain the confidence score $c^m_{ti}$ of each human joint $h^m_{ti}$, where $i=1,...,N$ and $N$ is the number of joints.
% $F_{tm}=\{v_{ti}|t=1,...,T, i=1,...,N\}$.
Then, we select the frame with the highest sum of confidence scores in the estimated joints within a segment because the higher the sum is, the more accurately the human joints are detected.
Finally, the sequence of the sampled frames is used directly to construct the human pose graph on both the spatial and temporal domains.
The node set of our skeletal graph can be written as
\begin{equation}
\begin{aligned}
\label{eq:informative}
V=\{v_{ti}|v_{ti}=h^{m^{*}}_{ti},\; m^*=\argmax_{m} {\overset{N}{\underset{i}{\sum}}c^m_{ti}}\}.
\end{aligned}
\end{equation}
To demonstrate that our frame selection algorithm performs better than the existing sampling strategy, we conduct an ablation study in Section~\ref{sec:ablation}.
As a result, OHA-GCN has a relatively complete form of graph that includes reliable joints.

% ------------------------------------------------------------------------
% \subsection{Combining human pose graph and object-related human pose graph convolutions}

\section{Experiments}

\subsection{Datasets}

\begin{table*}[t]\centering
\caption{Ablation study of the proposed framework over the IRD and ICVL-4 datasets. The baseline of our approach is ST-GCN~\cite{stgcn2018aaai} which is the state of the art for the skeleton-based action recognition using GCN.} 
\vspace{3mm}
\centering
\begin{tabular}{p{10.5cm}|cc} % {p{3cm}|cc|cc}

\hline
Method & IRD~($\%$) & ICVL-4~($\%$)   \\\hline
ST-GCN~\cite{stgcn2018aaai} (HP stream $+$ uniform samples) & 74.03 & 80.23   \\ \hline
OHA-GCN~(OHP-body stream $+$ informative samples) & 71.27  & 72.09   \\ 
OHA-GCN~(OHP-limbs stream $+$ informative samples) &  73.48 & 76.74  \\ 
OHA-GCN~(OHP-hands stream $+$ uniform samples) & 74.59  & 79.06  \\ 
OHA-GCN~(OHP-hands stream $+$ informative samples) &  76.24 & 81.40  \\ 
OHA-GCN~(HP stream $+$ informative samples) & 79.56 & 88.37  \\
 \hline 
OHA-GCN~(Two stream; HP $+$ OHP-hands $+$ informative samples) & \textbf{80.11} & \textbf{91.86}   \\ 
\hline
\end{tabular}
\\{HP: Human Pose, $~~$ OHP: Object-related Human Pose}
\label{tabular:all_res}
\end{table*}

The existing datasets~\cite{hu2015jointly,shahroudy2016ntu,xia2012view} are not suitable for use in evaluating our framework to recognize the object-related actions in real-world scenes %~\cite{hu2015jointly,shahroudy2016ntu,xia2012view}} 
for two reasons.
First, the existing datasets for skeleton-based action recognition were captured in constrained environments.
These datasets acquired from sensors are limited to indoor environments.
Therefore, the constrained datasets are insufficient for modeling realistic scenes because these datasets cover only a few variations in dynamic environments.
Second, these datasets usually include one actor in each image, which is not reflective of the scenes from real-world surveillance.
In our evaluation, instead of using the existing datasets, we use two realistic datasets captured in unconstrained real-world environments, which contain only raw video clips without skeleton data. The two datasets are described below.

\textbf{IRD dataset:} 
As for our own dataset, we have constructed an illegal rubbish dumping~(IRD) dataset that contains videos from CCTV cameras.
The dataset is used for the purpose of the general surveillance, especially monitoring illegal dumping.
The original videos are about 10 minutes long, and the resolution of the original video is 1280$\times$720 pixels~(HD).
We made, however, several video clips for annotations from the original videos by extracting the region of the event.
Each video clip, averaging 690 frames in length, is divided into two classes: $1)$ a garbage dumping action and $2)$ a normal event.
Humans tagged the ground truth, such as the person, the object the person carries, the start time of the action of dumping trash, and the end time.
Unlike other existing datasets for skeleton-based action recognition, this dataset is not limited to indoor environments and captured in unconstrained environments.
% Many actors appear in one scene, and 
There is a wide variation of conditions in dynamic environments, including viewpoint changes, illumination variance, and occlusions.
Most of all, this dataset is suitable for evaluating our method for recognizing object-related human actions.
We made a total of 1374 video clips and use 1193 of them for training and 181 for validation.

\textbf{ICVL-4 dataset:} 
The ICVL dataset~\cite{Jin:2018hda} includes dynamic sub-actions of multiple people at multiple locations in the surveillance scenes.
The actions are divided into 13 categories. 
%: sitting, standing, lying, stationary, walking, running, bicycling, nothing, texting, smoking, phoning, dumping, and others. 
The dataset contains 497 original videos, and one actor who is one of the people appearing in the video is marked by three sub-actions. 
For example, one person in the surveillance scene is standing, walking, and smoking.
The original video is divided into several video clips containing a single actor.
We use only these specific classes related to object-related actions, such as phoning, smoking, texting, and dumping, which will be called ICVL-4 in this paper.
The reason we choose the specific classes is that we have to verify the effectiveness of our method in recognizing object-related human actions. 
% We do not perform the action recognition where actions is represented by the mid level of human activities~\cite{moeslund2006survey}, such as running, walking and sitting. 
The evaluation using the entire class is meaningless, as the proposed strategies are the only major differences between previous skeleton-based approach and OHA-GCN.
% there are no big differences between previous skeleton-based approach and OHA-GCN except for the proposed strategies.
Therefore, we only conduct experiments with only the data from 4 classes, focusing on the object-related actions.
% where the high level of human activities is included.
A training set of 805 video clips and a validation set of 86 video clips are used for our experiments.
% \km{4개 클래스 썼다고 하고 넘어가도 될 것 같아서, mid-level 및 high-level 이야기는 일단 주석}

Both the IRD and ICVL-4 datasets provide only RGB video clips and action labels.
As our approach for action recognition is based on the skeleton, we use the pose estimation algorithm~\cite{cao2017realtime} and our detection algorithm.
Through the algorithms, 25 human joints are extracted, and the position of an object is obtained. % by Section~\ref{sec:obj-detect}.
There is a total of 26 nodes including both human joints and an object.
We make each video clip as a~$(3 ,T ,26)$ tensor, where the first dimension of the tensor represents 3 channels for the 2D coordinates~$(x, y)$ and confidence scores.
If more than two people exist in one video clip, we select the person with the highest sum of confidence scores as the actor for the recognition.
$T$ is the length of the frames, and 26 is the number of nodes here.
We use the top-1 accuracy, which is a conventional evaluation method for both datasets.
% 
%-------------------------------------------------------------------------
% \subsection{Implementation details}
% Our two-stream GCN designed on spatial and temporal domain employs both graphs for action recognition, selectively discarding ambiguous frames for construction of graphs consisting of relatively accurate joints.
% We argue that , since skeletal data are not always available in real-world applications. 
% 기존의 ST-GCN은 frame수가 다른 비디오로부터 뽑은 포즈정보를 그래프로 만들어 입력으로 넣었다. 하지만 frame수의 따라 temporal한 joint connection인 edge의 개수도 달라지므로, 그래프 구조 자체에도 변화가 생기는 것이다. 따라서 일정한 frame수의 skeleton sequence로 만들어주는 과정을 거친다. 이 때, key frame selection으로 중요한 포즈 정보를 담은 프레임들을 찾아볼 것이다.
%-------------------------------------------------------------------------
\subsection{Experimental result}
\subsubsection{Ablation study}
\label{sec:ablation}
\quad In the ablation study, we first investigate the role of 
the proposed strategy of selecting informative frames for skeleton-based action recognition. Second, we examine the strategy for how to connect the object to the human body in OHA-GCN in formulating object-related pose graphs.
Finally, we verify the effectiveness of the object-related human pose in recognizing object-related actions. 
We will show improvement using the proposed components in our framework compared with the state-of-the-art method, ST-GCN~\cite{stgcn2018aaai}, for skeleton-based action recognition.

\textbf{Frame selection strategies.} 
We compare the strategies of selecting frames on the IRD and ICVL-4 datasets: the proposed informative frame sampling and the existing uniform sampling.
Informative samples are selected from the informative frame selection strategy, and uniform samples are obtained from the uniform sampling strategy used in ST-GCN ~\cite{stgcn2018aaai}.
The uniform sampling selects $T$ frames at the same interval from a video.
% We demonstrate that the informative frame selection enables us to construct relatively complete graphs by removing redundant pose information. 
The comparisons between uniform and informative samples are performed using the human pose~(HP) stream and the object-related human pose with hands connection~(OHP-hands) stream.
Our results, as observed in Table~\ref{tabular:all_res}, show that the informative frame selection is a better strategy than uniform sampling. 
In particular, the informative frame selection in HP stream achieves a 5.5\%~(on IRD) and a 8\%~(on ICVL-4)  improvement compared to ST-GCN using uniform samples.
We argue that the improved performance comes from building more meaningful graphs by ignoring redundant information and focusing on accurate poses.
Our methods use the informative frame selection strategy for the following experiments.

\textbf{Graph construction strategies.} 
Approaches to connecting the object to the human body are explored empirically. `OHP-body' indicates connections between all the human joints and the object; `OHP-limbs' indicates connections between the human joints belonging to arms or legs and the object; `OHP-hands' denotes connections between the human joints of both hands and the object.
`OHP-body' shows lower performance than ST-GCN because the graph structure becomes complex and introduces redundant information.
It means that the more abstracted form the object-related pose graph takes, the better its performance.
As shown in Table~\ref{tabular:all_res}, `OHP-hands' shows the best performance among all the strategies.
We believe this result is reasonably good because object-related human actions usually occur in human hands. 
Consequently, we use the connections between human hands and an object in other experiments.
%We leave other methods of the graph construction to future works.

\textbf{HP stream and OHP stream.} 
We also compare the human pose stream with the object-related human pose stream in the Table~\ref{tabular:all_res}. 
The method using the object-related poses with the uniform sampling strategy~(OHP-hands stream $+$ uniform samples) does not improve performance compared to the human poses~(HP stream) with the same strategy~(ST-GCN; HP stream $+$ uniform samples).
While, 
The usage of object-related poses with the informative frame selection~(OHP-hands stream $+$ informative samples) achieves a 2.2\%~(on IRD) and a 1.2\%~(on ICVL-4) improvements compared to ST-GCN. The improvement is minimal, however, in the case of `HP stream $+$ informative samples'.
This difference in the performances is the result of the fact that the frame selection strategy is more informative to the human pose graph than the object-related one. By combining the HP and OHP streams, the two-stream OHA-GCN 
achieves the best performance of a 80.11\% and a 91.86\% on the IRD and ICVL-4 datasets, respectively, which 
is a significant improvement over the state-of-the-art, ST-GCN for skeleton-based action recognition.

\subsubsection{Comparison with CNN methods using images}
\quad Our method is compared with other CNN-based methods, evaluated on the ICVL-4 datasets.
Baseline CNNs are models of RGB modality, using only RGB images as input for the network.
In fact, this is not a fair comparison because our approach is based on pose modality while CNN methods are based on RGB modality.
Still, we want to show that even without the appearance information that is high dimensionality, our method can achieve a comparable performance.
% Our method is also compared with other baselines based on CNNs.
We establish baseline models whose architectures are LeNet~\cite{lecun1998gradient}, VGG16~\cite{simonyan2014very}, and ResNet50~\cite{he2016deep}, respectively.
% , AlexNet~\cite{krizhevsky2012imagenet}
As shown in Table~\ref{tabular:res_icvl}, the accuracy of our method is lower than some of CNN-based methods, as expected. 
% We argue that these results are RGB-based baselines are  although they achieves better performance.
This result is explained by the difference in the modalities.
RGB modality has more information for actions while pose-based representation is a compressed representation of the human body. 
Still, we argue that our pose-based approach achieves not only fairly high performance but also faster speed than RGB-based models.
% Since we not only use the pose-based approach, but also discard redundant frames through informative frame strategy.
To examine this argument, we perform speed comparisons with the RGB-based methods.
The test is performed on GTX TITAN X GPU, and other details of our implemented environments are described in Section~\ref{application}.
Using the ICVL-4 datasets, we measure the average runtime that the networks take to process one frame.
Our method can operate at around 3571 frames per second~(FPS), much faster speed than the CNN-based methods.
% Our method can operate at much more fast speed than the CNN methods, taking around 3571 frames per second~(FPS).
The LeNet works relatively quickly with light computation, but its accuracy is inferior to that of our method.
The ResNet achieves better results than our methods, but cannot operate quickly due to heavy computation.
% In our view, the computation time is reduced successfully mainly due to pose-based representation and our frame selection strategy.
Given these speed/accuracy trade-offs, our method has more advantages than other methods because it achieves the comparable results at high speed even without the RGB modality.

%%%%%%%%%%%%%%%%%%%%%%%%%%%%%%%%%%
\begin{table}
\caption{Accuracy and speed comparisons with the CNN-based methods using RGB image inputs on the ICVL-4 dataset.}
\vspace{3mm}
\centering
\begin{tabular}{p{4cm}|cc} % {p{3cm}|cc|cc}

\hline 

Method & Acc.($\%$) & Speed~($ms$)  \\ \hline 

CNN (LeNet~\cite{lecun1998gradient}) &  87.21 &  3.66 \\ 
CNN (VGG16~\cite{simonyan2014very}) &  95.35 &  24.56  \\ 
CNN (ResNet50~\cite{he2016deep}) &  \textbf{96.51} & 17.26 \\  \hline
ST-GCN~\cite{stgcn2018aaai} & 80.23 & 0.1321 \\ \hline
OHA-GCN (HP stream) & 88.37 & \textbf{0.1306} \\ 
OHA-GCN (OHP stream) & 81.40 & 0.1319 \\ 
OHA-GCN (Two Stream) & 91.86  & 0.2447 \\ \hline 
\end{tabular}
{HP: Human Pose,$~~$ OHP: Object-related Human Pose}
\label{tabular:res_icvl}
\end{table}

%%%%%%%%%%%%%%%%%%%%%%%%%%%%%%%%%%%
%\input{table/res_speed.tex}

\subsection{Applications}
\label{application}

Not only do our methods show promising results, but they also open up the possibility of many applications, building a real-world surveillance system. 
As a number of people appear throughout the original video in both datasets, there are many actors and actions to be recognized.
To prevent multiple people from obstructing an action of one actor in a video, we apply a tracking algorithm to extract the regions where a single person exists.
For the tracking, we adopt the online tracking-by-detection method based on the Hungarian matching algorithm~\cite{Yoo:2016cf} using pose information.
If more than two people overlap in one patch obtained from tracking, we use the person whose joints have the highest sum of confidence scores as an actor for action recognition.

To examine the real-time performance of our proposed method in the surveillance systems, we analyze the runtime of the overall pipeline.
Our method has been implemented with Pytorch deep learning framework, and the runtime is measured on GTX TITAN X GPU and Intel Core i7-6700K 4.0Ghz.
For applications in real-world surveillance, we also use tracking algorithms whose runtime is 12.87 ms using only CPU. % 94.5 ms
In our evaluation on the IRD dataset, foreground detection~\cite{cao2017realtime} and object detection operate at up to 45.78 ms and 14.93 ms using only CPU, respectively.
The frame size in the original videos is 1920$\times$1080, but we resize it to 656$\times$368 which is the size recommended by $OpenPose$~\cite{cao2017realtime}.
The pose estimation algorithm takes 74.62 ms using one GPU, 38.02 ms using two GPUs, and 20.92 ms using three GPUs.
The overall pipeline, including the runtime of OHA-GCN, takes around 94.78 ms~(10.6 FPS) on average, which can be applied to real-time monitoring.
The experimental results and runtime analysis imply that our approach can be successfully implemented in real-world scenarios, achieving real-time performance.

\section{Conclusion}
Our main contributions are summarized as follows. 
First, we propose a new framework, object-related human action recognition~(OHA-GCN), which uses the graphs of human poses and the object-related human poses to understand object-related human actions.
Second, to overcome the difficulties of skeleton-based action recognition in real-world scenarios, we explore good strategies, including informative frame selection and construction of the object-related human poses. 
Third, we introduce a new dataset of illegal rubbish dumping~(IRD), which compiles realistic data in unconstrained environments. 
Fourth, our framework can run in real time while achieving significantly better performance than the state-of-the-art algorithm for skeleton-based action recognition.
It is noted that the proposed strategies offer meaningful insights into how to incorporate contextual information into human-object interaction recognition in the future.

%%%%% Acknowledgement. After accept.
% \textbf{Acknowledgement.}  
% This work was partly supported by the ICT R\&D program of MSIP/IITP (No.B0101-15-0552, Development of Predictive Visual Intelligence Technology and No.B0101-15-0266, Development of High Performance Visual BigData Discovery Platform) and Brain Korea 21 Plus Project.

\vspace{2mm}
% \small
\noindent\textbf{Acknowledgements}
\ This work was partly supported by  
Next-Generation ICD Program through NRF funded by Ministry of S\&ICT [2017M3C4A7077582], 
the ICT R\&D program of MSIT/IITP. (No.B0101-15-0266, Development of High Performance Visual BigData Discovery Platform for Large-Scale Realtime Data Analysis and No.B0101-15-0552, Predictive Visual Intelligence Technology).
\normalsize

{\small
\bibliographystyle{ieee}
\bibliography{egbib}

\begin{thebibliography}{10}\itemsep=-1pt

\bibitem{Barnich:2011ks}
O.~Barnich and M.~Van~Droogenbroeck.
\newblock {ViBe: A Universal Background Subtraction Algorithm for Video
  Sequences}.
\newblock {\em IEEE Transactions on Image Processing}, 20(6):1709--1724, June
  2011.

\bibitem{bruna2013spectral}
J.~Bruna, W.~Zaremba, A.~Szlam, and Y.~LeCun.
\newblock Spectral networks and locally connected networks on graphs.
\newblock In {\em ICLR}, 2014.

\bibitem{cao2017realtime}
Z.~Cao, T.~Simon, S.-E. Wei, and Y.~Sheikh.
\newblock Realtime multi-person 2d pose estimation using part affinity fields.
\newblock In {\em CVPR}, 2017.

\bibitem{carreira2017quo}
J.~Carreira and A.~Zisserman.
\newblock Quo vadis, action recognition? a new model and the kinetics dataset.
\newblock In {\em CVPR}, 2017.

\bibitem{dalal2005histograms}
N.~Dalal and B.~Triggs.
\newblock Histograms of oriented gradients for human detection.
\newblock In {\em CVPR}, 2005.

\bibitem{defferrard2016convolutional}
M.~Defferrard, X.~Bresson, and P.~Vandergheynst.
\newblock Convolutional neural networks on graphs with fast localized spectral
  filtering.
\newblock In {\em NIPS}, 2016.

\bibitem{duvenaud2015convolutional}
D.~K. Duvenaud, D.~Maclaurin, J.~Iparraguirre, R.~Bombarell, T.~Hirzel,
  A.~Aspuru-Guzik, and R.~P. Adams.
\newblock Convolutional networks on graphs for learning molecular fingerprints.
\newblock In {\em NIPS}, 2015.

\bibitem{feichtenhofer2016convolutional}
C.~Feichtenhofer, A.~Pinz, and A.~Zisserman.
\newblock Convolutional two-stream network fusion for video action recognition.
\newblock In {\em CVPR}, 2016.

\bibitem{fernando2015modeling}
B.~Fernando, E.~Gavves, J.~M. Oramas, A.~Ghodrati, and T.~Tuytelaars.
\newblock Modeling video evolution for action recognition.
\newblock In {\em CVPR}, 2015.

\bibitem{Gkioxari:2018uv}
G.~Gkioxari, R.~Girshick, P.~Dollar, and K.~He.
\newblock {Detecting and Recognizing Human-Object Interactions}.
\newblock In {\em CVPR}, 2018.

\bibitem{he2016deep}
K.~He, X.~Zhang, S.~Ren, and J.~Sun.
\newblock Deep residual learning for image recognition.
\newblock In {\em CVPR}, 2016.

\bibitem{Hu:2018vj}
H.~Hu, J.~Gu, Z.~Zhang, J.~Dai, and Y.~Wei.
\newblock {Relation Networks for Object Detection}.
\newblock In {\em CVPR}, 2018.

\bibitem{hu2015jointly}
J.-F. Hu, W.-S. Zheng, J.~Lai, and J.~Zhang.
\newblock Jointly learning heterogeneous features for rgb-d activity
  recognition.
\newblock In {\em CVPR}, 2015.

\bibitem{ji20133d}
S.~Ji, W.~Xu, M.~Yang, and K.~Yu.
\newblock 3d convolutional neural networks for human action recognition.
\newblock {\em IEEE transactions on pattern analysis and machine intelligence},
  35(1):221--231, 2013.

\bibitem{Jin:2018hda}
C.-B. Jin, T.~D. Do, M.~Liu, and H.~Kim.
\newblock {Real-Time Action Detection in Video Surveillance using a Sub-Action
  Descriptor with Multi-Convolutional Neural Networks}.
\newblock {\em Journal of Institute of Control, Robotics and Systems},
  24(3):298--308, Mar. 2018.

\bibitem{ke2017new}
Q.~Ke, M.~Bennamoun, S.~An, F.~Sohel, and F.~Boussaid.
\newblock A new representation of skeleton sequences for 3d action recognition.
\newblock In {\em CVPR}, 2017.

\bibitem{Kim:2012bo}
S.~W. Kim, K.~Yun, K.~M. Yi, S.~J. Kim, and J.~Y. Choi.
\newblock {Detection of moving objects with a moving camera using non-panoramic
  background model}.
\newblock {\em Machine Vision and Applications}, 24(5):1015--1028, Oct. 2012.

\bibitem{kim2017interpretable}
T.~S. Kim and A.~Reiter.
\newblock Interpretable 3d human action analysis with temporal convolutional
  networks.
\newblock In {\em CVPR Workshops}, 2017.

\bibitem{kipf2016semi}
T.~N. Kipf and M.~Welling.
\newblock Semi-supervised classification with graph convolutional networks.
\newblock In {\em ICLR}, 2017.

\bibitem{laptev2008learning}
I.~Laptev, M.~Marszalek, C.~Schmid, and B.~Rozenfeld.
\newblock Learning realistic human actions from movies.
\newblock In {\em CVPR}, 2008.

\bibitem{lecun1998gradient}
Y.~LeCun, L.~Bottou, Y.~Bengio, and P.~Haffner.
\newblock Gradient-based learning applied to document recognition.
\newblock {\em Proceedings of the IEEE}, 86(11):2278--2324, 1998.

\bibitem{li2015gated}
Y.~Li, D.~Tarlow, M.~Brockschmidt, and R.~Zemel.
\newblock Gated graph sequence neural networks.
\newblock In {\em ICLR}, 2016.

\bibitem{liu2016spatio}
J.~Liu, A.~Shahroudy, D.~Xu, and G.~Wang.
\newblock Spatio-temporal lstm with trust gates for 3d human action
  recognition.
\newblock In {\em ECCV}, 2016.

\bibitem{moeslund2006survey}
T.~B. Moeslund, A.~Hilton, and V.~Kr{\"u}ger.
\newblock A survey of advances in vision-based human motion capture and
  analysis.
\newblock {\em Computer vision and image understanding}, 104(2-3):90--126,
  2006.

\bibitem{newell2016stacked}
A.~Newell, K.~Yang, and J.~Deng.
\newblock Stacked hourglass networks for human pose estimation.
\newblock In {\em ECCV}, 2016.

\bibitem{niepert2016learning}
M.~Niepert, M.~Ahmed, and K.~Kutzkov.
\newblock Learning convolutional neural networks for graphs.
\newblock In {\em ICML}, 2016.

\bibitem{pavlakos2017coarse}
G.~Pavlakos, X.~Zhou, K.~G. Derpanis, and K.~Daniilidis.
\newblock Coarse-to-fine volumetric prediction for single-image 3d human pose.
\newblock In {\em CVPR}, 2017.

\bibitem{shahroudy2016ntu}
A.~Shahroudy, J.~Liu, T.-T. Ng, and G.~Wang.
\newblock Ntu rgb+ d: A large scale dataset for 3d human activity analysis.
\newblock In {\em CVPR}, 2016.

\bibitem{shotton2011real}
J.~Shotton, A.~Fitzgibbon, M.~Cook, T.~Sharp, M.~Finocchio, R.~Moore,
  A.~Kipman, and A.~Blake.
\newblock Real-time human pose recognition in parts from single depth images.
\newblock In {\em CVPR}, pages 1297--1304. Ieee, 2011.

\bibitem{simonyan2014two}
K.~Simonyan and A.~Zisserman.
\newblock Two-stream convolutional networks for action recognition in videos.
\newblock In {\em Advances in neural information processing systems}, pages
  568--576, 2014.

\bibitem{simonyan2014very}
K.~Simonyan and A.~Zisserman.
\newblock Very deep convolutional networks for large-scale image recognition.
\newblock {\em arXiv preprint arXiv:1409.1556}, 2014.

\bibitem{tang2018deep}
Y.~Tang, Y.~Tian, J.~Lu, P.~Li, and J.~Zhou.
\newblock Deep progressive reinforcement learning for skeleton-based action
  recognition.
\newblock In {\em CVPR}, 2018.

\bibitem{tran2015learning}
D.~Tran, L.~Bourdev, R.~Fergus, L.~Torresani, and M.~Paluri.
\newblock Learning spatiotemporal features with 3d convolutional networks.
\newblock In {\em Proceedings of the IEEE international conference on computer
  vision}, pages 4489--4497, 2015.

\bibitem{vemulapalli2014human}
R.~Vemulapalli, F.~Arrate, and R.~Chellappa.
\newblock Human action recognition by representing 3d skeletons as points in a
  lie group.
\newblock In {\em CVPR}, 2014.

\bibitem{wang2013action}
H.~Wang and C.~Schmid.
\newblock Action recognition with improved trajectories.
\newblock In {\em ICCV}, 2013.

\bibitem{wang2016temporal}
L.~Wang, Y.~Xiong, Z.~Wang, Y.~Qiao, D.~Lin, X.~Tang, and L.~Van~Gool.
\newblock Temporal segment networks: Towards good practices for deep action
  recognition.
\newblock In {\em ECCV}, 2016.

\bibitem{wang2016graph}
P.~Wang, C.~Yuan, W.~Hu, B.~Li, and Y.~Zhang.
\newblock Graph based skeleton motion representation and similarity measurement
  for action recognition.
\newblock In {\em ECCV}, 2016.

\bibitem{xia2012view}
L.~Xia, C.-C. Chen, and J.~K. Aggarwal.
\newblock View invariant human action recognition using histograms of 3d
  joints.
\newblock In {\em CVPR Workshops, 2012 IEEE computer society conference on},
  pages 20--27, 2012.

\bibitem{stgcn2018aaai}
S.~Yan, Y.~Xiong, and D.~Lin.
\newblock Spatial temporal graph convolutional networks for skeleton-based
  action recognition.
\newblock In {\em AAAI}, 2018.

\bibitem{yao2012coupled}
A.~Yao, J.~Gall, and L.~Van~Gool.
\newblock Coupled action recognition and pose estimation from multiple views.
\newblock {\em International journal of computer vision}, 100(1):16--37, 2012.

\bibitem{Yi:2013gp}
K.~M. Yi, K.~Yun, S.~W. Kim, H.~J. Chang, H.~Jeong, and J.~Y. Choi.
\newblock {Detection of Moving Objects with Non-stationary Cameras in 5.8ms:
  Bringing Motion Detection to Your Mobile Device}.
\newblock In {\em CVPR Workshops}, 2013.

\bibitem{Yoo:2016cf}
H.~Yoo, K.~Kim, M.~Byeon, Y.~Jeon, and J.~Y. Choi.
\newblock {Online Scheme for Multiple Camera Multiple Target Tracking Based on
  Multiple Hypothesis Tracking}.
\newblock {\em IEEE Transactions on Circuits and Systems for Video Technology},
  27(3):454--469, Mar. 2017.

\bibitem{Yun:2017jd}
K.~Yun, J.~Lim, and J.~Y. Choi.
\newblock {Scene conditional background update for moving object detection in a
  moving camera}.
\newblock {\em Pattern Recognition Letters}, 88:57--63, 2017.

\bibitem{zhang2017geometric}
S.~Zhang, X.~Liu, and J.~Xiao.
\newblock On geometric features for skeleton-based action recognition using
  multilayer lstm networks.
\newblock In {\em WACV}, 2017.

\end{thebibliography}
}

\end{document}